# Dual Pointer Network for Fast Extraction of Multiple Relations in a Sentence

**Seongsik Park [1] and Harksoo Kim [2,***

[1] Kangwon National University; a163912@kangwon.ac.kr
[2] Konkuk University; nlpdrkim@konkuk.ac.kr
**\*** Correspondence: nlpdrkim@konkuk.ac.kr; Tel.: +82-2-450-3499

**Featured Application: Ontology construction module for AI applications**

**Abstract:** Relation extraction is a type of information extraction task that recognizes semantic relationships between entities in a sentence. Many previous studies have focused on extracting only one semantic relation between two entities in a single sentence. However, multiple entities in a sentence are associated through various relations. To address this issue, we propose a relation extraction model based on a dual pointer network with a multi-head attention mechanism. The proposed model finds *n*-to-*1* subject–object relations using a forward object decoder. Then, it finds *1*-to-*n* subject–object relations using a backward subject decoder. Our experiments confirmed that the proposed model outperformed previous models, with an F1-score of 80.8% for the ACE-2005 corpus and an F1-score of 78.3% for the NYT corpus.

**Keywords:** Relation extraction; dual pointer network; context-to-entity attention

## 1. Introduction

Relation extraction is a task that involves recognizing semantic relations (i.e., tuple structures; [subject, relation, object triples]) among entities in a sentence [1]. Zeng et al. [2] divided sentences into three types according to the triplet overlap degree: normal, entity pair overlap (EPO), and single entity overlap (SEO). In the normal type, the triples do not have overlapped entities; in the EPO type, some triples have an overlapped entity pair; and in the SEO type, some triplets have an overlapped entity, but these triplets do not have overlapped entity pairs. In this study, we focus on promptly extracting both the normal and SEO types because most relations are included in these types, as shown in Figure 1.

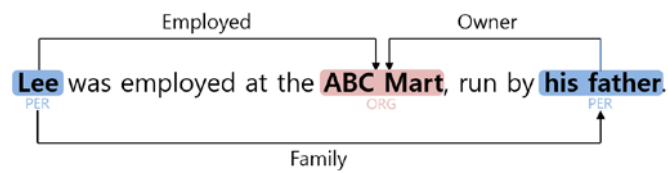

**Figure 1.** Subject-relation-object triples in a sentence.

In Figure 1, *[Lee, employed, ABC Mart]*, *[Lee, Family, his Father]* and *[his Father, Owner, ABC Mart]* are SEO types. To promptly extract these relations, we adopt the concept of dependency parsing in which dependent words point to the head words by scanning each word in a sentence. We propose a

---

dual pointer network model to efficiently extract multiple relations from a sentence through forward scanning (i.e., scanning from the first word to the last) and backward scanning (i.e., scanning from the last word to the first). The proposed model discovers an object of the current subject during forward scanning. Through forward scanning, all normal type relations can be found. However, SEO type relations are only partially found because a subject should point to only one object in the pointer network architecture. To address this limitation, the proposed model performs backward scanning to identify a subject of the current object.

The remainder of this paper is organized as follows. In Section 2, we review previous studies on relation extraction. Section 3 describes the proposed dual pointer network model. In Section 4, we elaborate on the experimental setup and results. Finally, we conclude the study in Section 5.

## 2. Previous Works

With the significant success of deep neural networks in the field of natural language processing, many researchers have proposed various relation extraction models based on convolutional neural networks (CNNs). These include the CNN model based on max-pooling [3], the CNN model based on multiple sizes of kernels [4], the combined CNN model [5], and the contextualized graph convolutional network (C-GCN) model [6]. Relation extraction models based on recurrent neural network (RNNs) have also been proposed, including the long-short term memory (LSTM) model based on the dependency tree [7] and the LSTM model using the position-aware attention technique [8]. These models have focused on normal type extraction (i.e., extracting only one relation between two entities from a single sentence). However, many entities in a single sentence can form multiple relations. To resolve this problem, some studies have proposed multiple relation extraction. For example, Luan et al. [9] treated triples in sentences as a graph and proposed a multiple relations extraction model that iteratively extracts spans between triples in the graph. In the present study, we propose a relation extraction model to simultaneously find all possible relations among multiple entities in a sentence. The proposed model is based on the pointer network [10]. The pointer network is a sequence-to-sequence (Seq2Seq) model in which an attention mechanism [11] is modified to learn the conditional probability of an output, whose values correspond to positions in a given input sequence. We modify the pointer network to include dual decoders, an object decoder (a forward decoder) and a subject decoder (a backward decoder). The object decoder extracts *n*-to-*1* relations as shown in the following example: *[Lee, employed, ABC Mart]* and *[his Father, Owner, ABC Mart]* are extracted from the sentence. The subject decoder extracts *1*-to-*n* relations as shown in the following example: *[Lee, employed, ABC Mart]* and *[Lee, Family, his Father]* are extracted from the sentence.

## 3. Dual Pointer Network Model for Relation Extraction

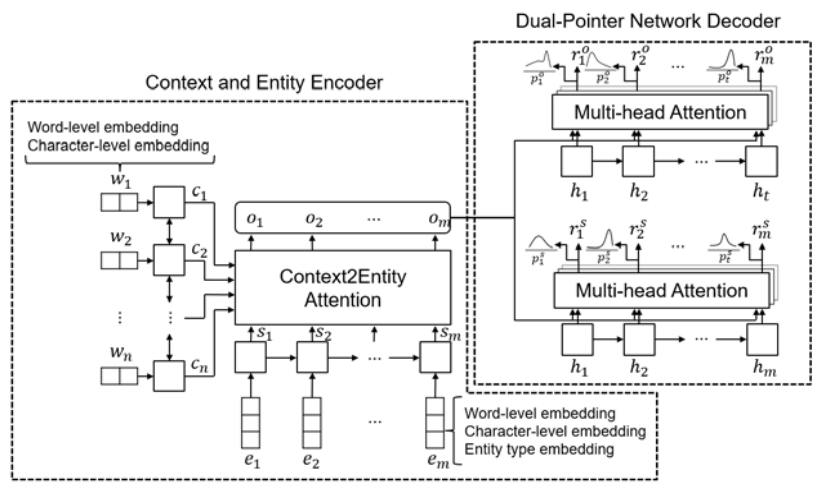

**Figure 2.** Overall architecture of dual pointer networks for relation extraction.

Figure 2 illustrates the architecture of the proposed model. This consists of two parts, a context and entity encoder, and a dual pointer network decoder.

*3.1. Context and Entity Encoder*

The context and entity encoder computes the degree of association between words and entities in a given sentence. For example, $\{w_1, w_2, ..., w_i\}$ and $\{e_1, e_2, ..., e_m\}$ refer to word and entity embedding vectors, respectively. Figure 3 illustrates the process of word and entity embedding.

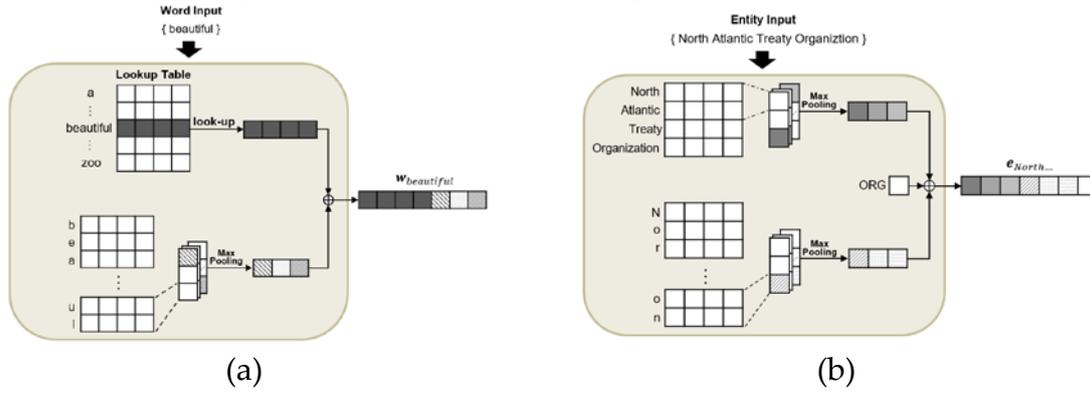

**Figure 3.** (a) Word embedding process (b) Entity embedding process.

As shown in Figure 3, the word embedding vectors are concatenations of two types of embeddings: word-level GloVe [12] embeddings for representing the meaning of words and character-level CNN embeddings [13] for addressing out-of-vocabulary problems. The entity embedding vectors are concatenations of three types of embeddings: word-level CNN embedding for representing the meaning of entities composed of multiple words, character-level CNN embedding for addressing out-of-vocabulary problems, and entity type embedding for representing the categorical information of input entities. Each word in the word-level CNN embedding is represented by word-level GloVe embeddings. The word embedding vectors are used as input for a bidirectional LSTM network to obtain contextual information as follows:

$$\vec{c}_i = \text{LSTM}(w_i, \vec{c}_{i-1}),$$
$$\overleftarrow{c}_i = \text{LSTM}(w_i, \overleftarrow{c}_{i-1}), \quad (1)$$
$$c_i = [\vec{c}_i; \overleftarrow{c}_i],$$

where $w_i$ is an embedding vector of the *i*-th word in a sentence, and $[\vec{c}_i; \overleftarrow{c}_i]$ is a concatenation of $\vec{c}_i$ and $\overleftarrow{c}_i$ that represents the output vectors of a forward LSTM and a backward LSTM, respectively. The entity embedding vectors are used as input for a forward LSTM network because the entities are listed in the order they appear in a sentence, as shown below.

$$s_t = \text{LSTM}(e_t, s_{t-1}), \quad (2)$$

where $e_t$ is an embedding vector of the *t*-th one among all entities occurring in a sentence, and $s_t$ is an output vector encoded by a forward LSTM. The output vectors of the bidirectional LSTM network $\{c_1, c_2, ..., c_i\}$ and the forward LSTM network $\{s_1, s_2, ..., s_t\}$ are used as input for the context-to-entity attention layer (as shown in Figure 2), to compute the relative degrees of association between words and entities. This is similar to the well-known multi-head attention mechanism [14] as shown below.

$$q_j = w^a * split(q, n)_j,$$
$$a_j = softmax(\frac{q_j k_j}{\sqrt{d}}),$$
$$head_j = a_j v_j, \quad (3)$$
$$o_t = \text{relu}(w^o[head_0; head_1; head_2; ...; head_n] + b^o),$$

where the query $q$ is set to $s_t$, the key $k$ and the value $v$ are set to C's. The query $q$ is split into $n$ vectors, where $n$ is the number of heads. The attention score $a_j$ is calculated by a scaled-dot product, where $d$ is a normalization factor. The context-to-entity layer output $o_t$ is determined through a fully-connected neural network (FNN) using a concatenation of $n$ heads as input.

*3.2. Dual Pointer Network Decoder*

In a pointer network, attentions show the position distributions of an encoding layer. Because an attention is highlighted at only one position, the pointer network has a structural limitation when one entity forms relations with several entities (for instance, "Lee" in Figure 1). The proposed model adopts a dual pointer network decoder (see Figure 2) to overcome this limitation. The first decoder, called an object decoder, learns the position distribution from subjects to objects as follows:

$$h_t = [e_t; s_t],$$
$$g_t = \text{LSTM}(h_t, g_{t-1}),$$
$$score_t^{obj} = v^{obj} \tanh(w^{obj}[O; g_t]),$$
$$a_t^{obj} = \text{softmax}(score_t^{obj}),$$
$$\hat{p}_t^{obj} = \text{argmax}(a_t^{obj}),$$
$$\hat{r}_t^{obj} = \text{argmax}(u^{obj}\tanh(z^{obj}[a_t^{obj}O; g_t])),$$
(4)

where $h_t$ is a concatenation of the entity embedding vector $e_t$ and the LSTM-encoded entity embedding vector $s_t$, and the decoding vector $g_t$ (i.e., the *t*-th entity to determine its objects) is calculated by the forward LSTM. Then, $a_t^{obj}$ is the position distribution based on the attention scores $score_t^{obj}$ between $g_t$ and the other entities $o_1, \ldots, o_{t-1}, o_{t+1}, \ldots, o_m$ in the context-to-entity attention layer. $\hat{p}_t^{obj}$ and $\hat{r}_t^{obj}$ represent a position and a relation name of $g_t$'s object, respectively. The weighting parameters $v, w, u,$ and $z$ are set during the training phase. Conversely, the second decoder, called a subject decoder, learns the position distribution from objects to subjects in the same manner as the object decoder, as shown below.

$$score_t^{sub} = v^{sub} \tanh(w^{sub}[O; g_t]),$$
$$a_t^{sub} = \text{softmax}(score_t^{sub}),$$
$$\hat{p}_t^{sub} = \text{argmax}(a_t^{sub}),$$
$$\hat{r}_t^{sub} = \text{argmax}(u^{sub}\tanh(z^{sub}[a_t^{sub}O; g_t])),$$
(5)

where $\hat{p}_t^{sub}$ and $\hat{r}_t^{sub}$ represent a position and a relation name of $g_t$'s subject, respectively. In Figure 1, "Lee" should point to both "ABC mart" and "his father." This problem cannot be solved using the conventional forward decoder because it cannot point to multiple targets. However, the subject decoder (a backward decoder) resolves this problem, because "ABC mart" and "his father" can point to "Lee."

Additionally, we adopt a multi-head attention mechanism to improve the performance of the dual pointer network; this is shown in the following equation.

$$q_j = w^l * split(q, n)_j,$$
$$a_j = softmax(\frac{q_j k_j}{\sqrt{d}}),$$
$$head_j = a_j v_j,$$
$$\hat{p}_t = \text{argmax}(\frac{1}{n}\sum_{k=0}^{n} a_k),$$
$$\hat{r}_t = \text{argmax}(relu(w^r[head_0; head_1; head_2; \ldots; head_n] + b^r)),$$
(6)

where the query $q$ is set to $g_t$, the key $k$ and the value $v$ are set to $O$'s. The position distribution $\hat{p}_t$ is calculated by an average of $n$ multi-head attention vectors, and the relation name $\hat{r}_t$ is determined through an FNN using a concatenation of $n$ heads as the input.

*3.3. Implementation detail*

The context and entity encoder comprised 256 hidden units in each layer, and the dual pointer network decoder comprised 512 hidden units. We adopted a 0.1 drop-out probability for all the LSTM cells. We used 8 heads, with 32 units per head, for the multi-head attention. The vocabulary size and word-embedding size was set to 16,925 and 300, respectively. The filter size of the CNNs for character and word embeddings were 3, 4, and 5. The total number of filters was 100. 50-dimensional random initialized vectors were used for the character and entity embeddings. A cross-entropy function was used as a cost function to maximize the log-probability as follows:

$$CE(y,\tilde{y}) = -\sum_i^C y_i \log(\tilde{y}_i),$$
$$Loss = \frac{\alpha}{2}\left(CE(p^{sub},\tilde{p}^{sub}) + CE(p^{obj},\tilde{p}^{obj})\right) + \frac{(1-\alpha)}{2}(CE(r^{sub},\tilde{r}^{sub}) + CE(r^{obj},\tilde{r}^{obj})), \quad (7)$$

where $y$ is the target answer, $\tilde{y}$ is the score distribution of the model prediction, and $C$ is the number of target classes. The loss is calculated by the cross-entropy combination of all targets and predictions. The weighting factor α was experimentally set to 0.6 as a scalar value.

## 4. Evaluation

*4.1. Datasets and Experimental Setting*

We evaluated the proposed model using the following benchmark datasets.

**ACE-05 corpus [15]**: The ACE dataset includes seven major entity types and six major relation types. The ACE-05 corpus does not properly evaluate models that extract multiple triples from a sentence. Therefore, if some triples in the ACE-05 corpus share a sentence (*i.e.*, some triples occur in the same sentence), the triples were merged, as shown in Figure 4.

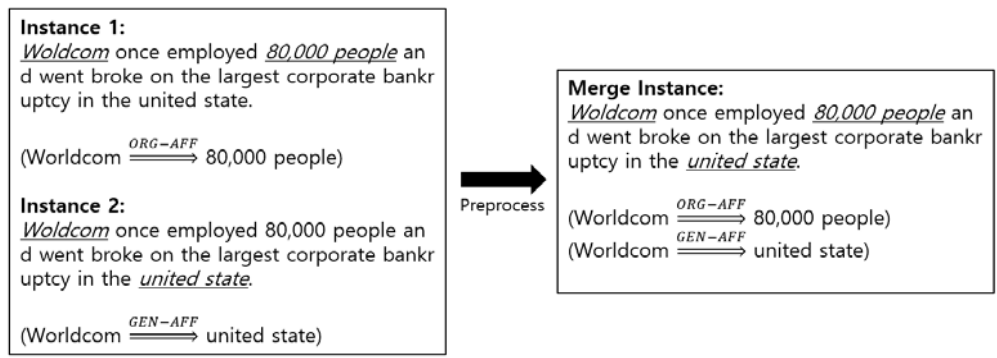

**Figure 4.** ACE-2005 data preprocess.

As a result, we obtained a dataset annotated with multiple triples. We divided the new dataset into a training set (5,023 sentences), a development set (629 sentences), and a test set (627 sentences) by a ratio of 8:1:1.

**NYT corpus [16]**. This is a news corpus sampled from news articles published in the NYT. The training data is automatically labeled using distant supervision. The NYT corpus was manually converted to a relation extraction dataset by Zheng et al. [17]. We excluded sentences without relation facts from Zheng's corpus. Finally, we obtained 66,202 sentences in total. We used 59,581 sentences for training and 6,621 for testing.

We adopted the standard micro precision, recall, and F1 score to evaluate the results:

$$\text{Recall} = \frac{\text{\# of correct predict}}{\text{\# of all triple in the dataset}}$$

$$\text{Precision} = \frac{\text{\# of correct predict}}{\text{\# of all triple in the model predict}} \quad (8)$$

$$F1 = \frac{2 * Pricision * Recall}{Precision + Recall}$$

*4.2. Experimental Results*

In the first experiment, we evaluated the effectiveness of the multi-head attention in the dual pointer network decoder; the results are summarized in Table 1. The evaluation was performed using the ACE-05 corpus.

**Table 1.** Performance for different attention mechanisms in the dual pointer network decoder

| Model | Recall | Precision | F1 |
| --- | --- | --- | --- |
| Single-head | 0.800 | 0.759 | 0.779 |
| Multi-head | 0.832 | 0.787 | 0.808 |

In Table 1, single-head refers to a conventional attention mechanism proposed by Bahdanau et al. [11]. As shown in Table 1, the multi-head attention mechanism used in the proposed model demonstrated better performance than the single-head one. Then, using the ACE-05 corpus, we evaluated the effectiveness of multi-head attention in the context and entity encoder; the results are summarized in Table 2.

**Table 2.** Performance for different attention mechanisms in the context and entity encoder

| Model | Recall | Precision | F1 |
| --- | --- | --- | --- |
| BIDAF-C2Q | 0.819 | 0.766 | 0.792 |
| BIDAF-C2Q&Q2C | 0.821 | **0.792** | 0.806 |
| Multi-head | 0.832 | 0.787 | 0.808 |

In Table 2, BIDAF [18] refers to a machine reading and comprehension (MRC) model based on a co-attention mechanism between a query and a context. C2Q and Q2C are refer to mean context-to-query attention and query-to-context attention used in the BIDAF model, respectively. As shown in Table 2, the multi-head attention mechanism used in the proposed model showed the best F1-score.

In the second experiment, we compared the proposed model with previous state-of-the-art models. Table 3 compares the performance of the proposed model and with other models for the ACE-2005 dataset.

**Table 3.** Performance comparison on ACE-2005 dataset

| Model | Recall | Precision | F1 |
| --- | --- | --- | --- |
| SPTree | 0.540 | 0.572 | 0.556 |
| FCM | 0.493 | 0.715 | 0.582 |
| DYGIE | 0.567 | 0.643 | 0.602 |
| Span-Level | 0.584 | 0.680 | 0.628 |
| Walk-Based | 0.595 | 0.697 | 0.642 |
| HRCNN | - | - | 0.741 |
| Our model | 0.832 | 0.787 | 0.808 |

In Table 3, SPTree [6] is a model that applies the dependency information between the entities. In FCM [19], handcrafted features are combined with word embeddings. DYGIE [9] dynamically generates spans between entities and spans' representations. Span-Level [20] jointly performs entity mention detection and relation extraction. HRCNN [21] is a hybrid model of CNN, RNN, and FNN. Walk-Based [22] is a graph-based neural network model. As shown in Table 1, the proposed model outperformed all models across all metrics. Table 4 compares the performance of the proposed model with existing models for the NYT Corpus.

Table 4. Performance comparisons on NYT corpus

| Model | Recall | Precision | F1 |
|---|---|---|---|
| NovelTag | 0.414 | 0.615 | 0.495 |
| MultiDecoder | 0.566 | 0.610 | 0.587 |
| GraphRE | 0.600 | 0.639 | 0.619 |
| Our model | 0.820 | 0.749 | 0.783 |

In Table 4, NovelTag [17] is an end-to-end model that extracts entities and their relations based on a novel tagging scheme designed for relation extraction. MultiDecoder [2] is a Seq2Seq-based model that combines the entity and relation extraction using a decoder with copy mechanism. GraphRE [23] is a joint model that extracts entities and their relation using graph convolutional networks (GCN) [24]. As shown in Table 4, the proposed model outperformed all models. It is not reasonable to directly compare the proposed model with these models because it requires gold-labeled entities, while the other models automatically extract entities from sentences. Although direct comparison is unfair, the proposed model exhibited considerably better performance. If we adopt a state-of-the-art named entity tagger based on BERT [25] with F1-scores of 0.9 or more, the proposed model is expected to show F1-scores of 0.662 or more based on simple multiplication.

The cases where the proposed model incorrectly extracted relations were also grouped in Table 5.

Table 5. Main reasons for errors (underline denotes incorrect results)

| Input sentence | Correct relation | Predicted relation |
|---|---|---|
| Iraqi forces responded with artillery fire | [Iraqi forces, Art, artillery] [Iraqi forces, Gen-aff, Iraqi] | [Iraqi forces, Part-whole, Iraqi] [Iraqi forces, Gen-aff, Iraqi] |
| It is the first time they have had freedom of movement with cars and weapons since the start of the intifada | [they, Art, cars] [they, Art, weapons] | [they, Art, cars] |
| It was in northern Iraq today that an eight artillery round hit the site occupied by Kurdish fighters near Chamchamal | [Kurdish fighters, Phys, the site] [the site, Phys, Chamchamal] [Kurdish, Gen-aff, Kurdish fighters] [the site, Part-whole, northern Iraq] | [Kurdish fighters, Phys, the site] [the site, Phys, Chamchamal] [the site, Part-whole, northern Iraq] [Kurdish fighters, Art, artillery] |

Most incorrect predictions included cases where the decoders incorrectly pointed out subjects or objects, and these incorrect entities lead to incorrect relation names, as shown in the first and third sentences in Table 5. In some cases, the decoder did not point out subjects or objects. As a result, any triples in a sentence were not omitted, as shown in the second sentence.

**5. Conclusion**

We proposed a relation extraction model to find all possible relations among multiple entities in a sentence simultaneously. The proposed model is based on pointer networks with multi-head attention mechanisms. To extract all possible relations from a sentence, we modified a single decoder into a dual decoder. In the dual decoder, the object decoder extracts *n*-to-*1* subject–object relations, and the subject decoder extracts *1*-to-*n* subject–object relations. The results from the experiments with the ACE-05 corpus and the NYT corpus confirmed that the proposed model shows an improvement in performance. Our future work will focus on an end-to-end model that directly extracts entities and their relations. In addition, we will focus on a method for improving performance using a large-scale language model like BERT [25].

**Author Contributions:** Conceptualization, H. Kim.; methodology, H. Kim.; software, S. Park.; validation, S. Park.; formal analysis, H. Kim.; investigation, H. Kim.; resources, S. Park.; data curation, S. Park.; writing—original draft preparation, S. Park.; writing—review and editing, H. Kim.; visualization, H. Kim.; supervision, H. Kim.; project administration, H. Kim.; funding acquisition, H. Kim.

**Funding:** This work was supported by the Institute for Information & communications Technology Planning & Evaluation(IITP) grant funded by the Korea government (MSIT) (No. 2013-0-00131, Development of Knowledge Evolutionary WiseQA Platform Technology for Human Knowledge Augmented Services).

**Acknowledgments:** We especially thank the members of the NLP laboratory at Kangwon National University and Konkuk University for their technical support.

**Conflicts of Interest:** The authors declare no conflict of interest.